\documentclass[a4paper,fleqn]{cas-dc}

\usepackage[numbers,sort&compress]{natbib}
\usepackage{amsmath,amssymb}
\usepackage{booktabs}
\usepackage{array}
\usepackage{placeins}
\usepackage{float}
\usepackage{graphicx}
\usepackage{microtype}
\usepackage{xspace}
\usepackage{xurl}

\usepackage{algorithm}
\usepackage{algpseudocode}
\algrenewcommand\algorithmicrequire{\textbf{Input:}}
\algrenewcommand\algorithmicensure{\textbf{Output:}}

\usepackage{multirow}
\usepackage{makecell}
\usepackage{tabularx}

\DeclareRobustCommand{\methodname}{Fed-MAPTrack\xspace}
\makeatletter
\let\fedmap@startsection\@startsection
\renewcommand\@startsection[6]{%
  \fedmap@startsection{#1}{#2}{#3}{#4}{#5}{#6\raggedright\hyphenpenalty=10000\exhyphenpenalty=10000\relax}}
\makeatother

\begin{document}
\let\WriteBookmarks\relax
\def\floatpagepagefraction{1}
\def\textpagefraction{.001}

\shorttitle{Fed-MAPTrack: Federated Modality-aware Prompt Tuning for RGB-X Tracking}
\shortauthors{Liu et al.}

\title[mode=title]{Fed-MAPTrack: Federated Modality-aware Prompt Tuning for RGB-X Tracking}

\author[1]{Xiangyu Liu}
\fnmark[1]

\author[1]{Yijie Shen}
\fnmark[1]

\author[2]{Yue Yin}

\author[3]{Zhitong Yin}

\author[4]{Yongqi Pan}
\cormark[1]

\author[1]{Wentao Fan}
\cormark[2]

\affiliation[1]{organization={Beijing Normal-Hong Kong Baptist University},
            postcode={519087},
            city={Zhuhai},
            state={Guangdong},
            country={China}}

\affiliation[2]{organization={Zhejiang University},
            postcode={310058},
            city={Hangzhou},
            state={Zhejiang},
            country={China}}

\affiliation[3]{organization={Peking University},
            postcode={100871},
            city={Beijing},
            country={China}}

\affiliation[4]{
    organization={Military Management College},
    addressline={National Defense University PLA China},
    postcode={100091},
    city={Beijing},
    country={China}
}

\cortext[cor1]{Corresponding author}
\cortext[cor2]{Corresponding author}
\fntext[fn1]{Xiangyu Liu and Yijie Shen contributed equally to this work.}

\begin{abstract}
RGB-X tracking complements RGB appearance with thermal, depth, or event observations, but centralized training requires collecting multimodal videos at a single location.
Federated training keeps raw videos local, but sensor heterogeneity can create a mismatch between parameter-sharing domains and the clients contributing to them.
This mismatch is particularly relevant when RGB-only clients coexist with sensor-equipped clients: their RGB observations remain useful, but they provide no auxiliary measurements for learning sensor-dependent interactions.
We propose \methodname, a modality-aware federated prompt tuning framework for RGB-X tracking.
The pretrained tracking backbone and prediction head are frozen, and clients optimize and exchange only prompt-related parameters.
The server aggregates RGB-shared parameters globally and maintains separate thermal, event, and depth banks for auxiliary-specific, fusion, and normalization parameters.
Selective participation retains RGB-only clients' contributions to the global RGB bank while excluding their fusion and normalization updates from modality-specific aggregation.
Across three matched training repetitions under a fixed client partition, \methodname improves the mean FedAvg scores by 2.35 AUC points on LasHeR, 3.45 SR points on VisEvent, and 2.11 F-score points on DepthTrack.
Among the nine evaluated methods, it achieves the highest mean thermal AUC and depth F-score, while its mean event SR is 0.07 points below that of Independent FL.
Controlled ablations show that extending the bucketing scope without a compatible participation rule can reduce performance.
These results highlight the need to jointly define the sharing domain and eligible contributors for each parameter group.
\end{abstract}

\begin{highlights}
\item Federated prompt tuning enables heterogeneous RGB-X tracking while keeping videos local.
\item RGB prompts are shared globally, while sensor-related parameters use modality banks.
\item Selective aggregation retains RGB updates from clients without auxiliary sensors.
\item Ablations show that modality bucketing requires compatible participation rules.
\end{highlights}

\maketitle

\section{Introduction}\label{sec:introduction}

Visual object tracking localizes a target throughout a video given its initial state.
Siamese and Transformer-based trackers have advanced template-search matching and target localization~\cite{siamfc,stark,ostrack}, but RGB appearance alone can be unreliable under low illumination, occlusion, motion blur, or background distraction.
RGB-X tracking supplements RGB with thermal, depth, or event observations.
These sensors provide complementary cues: thermal observations capture infrared appearance, depth describes scene geometry, and events encode brightness changes with high temporal resolution~\cite{lasher,depthtrack,visevent}.
Learning to exploit these complementary cues is essential for reliable tracking under challenging visual conditions.

\begin{figure}[pos=t]
  \centering
  \includegraphics[width=\columnwidth]{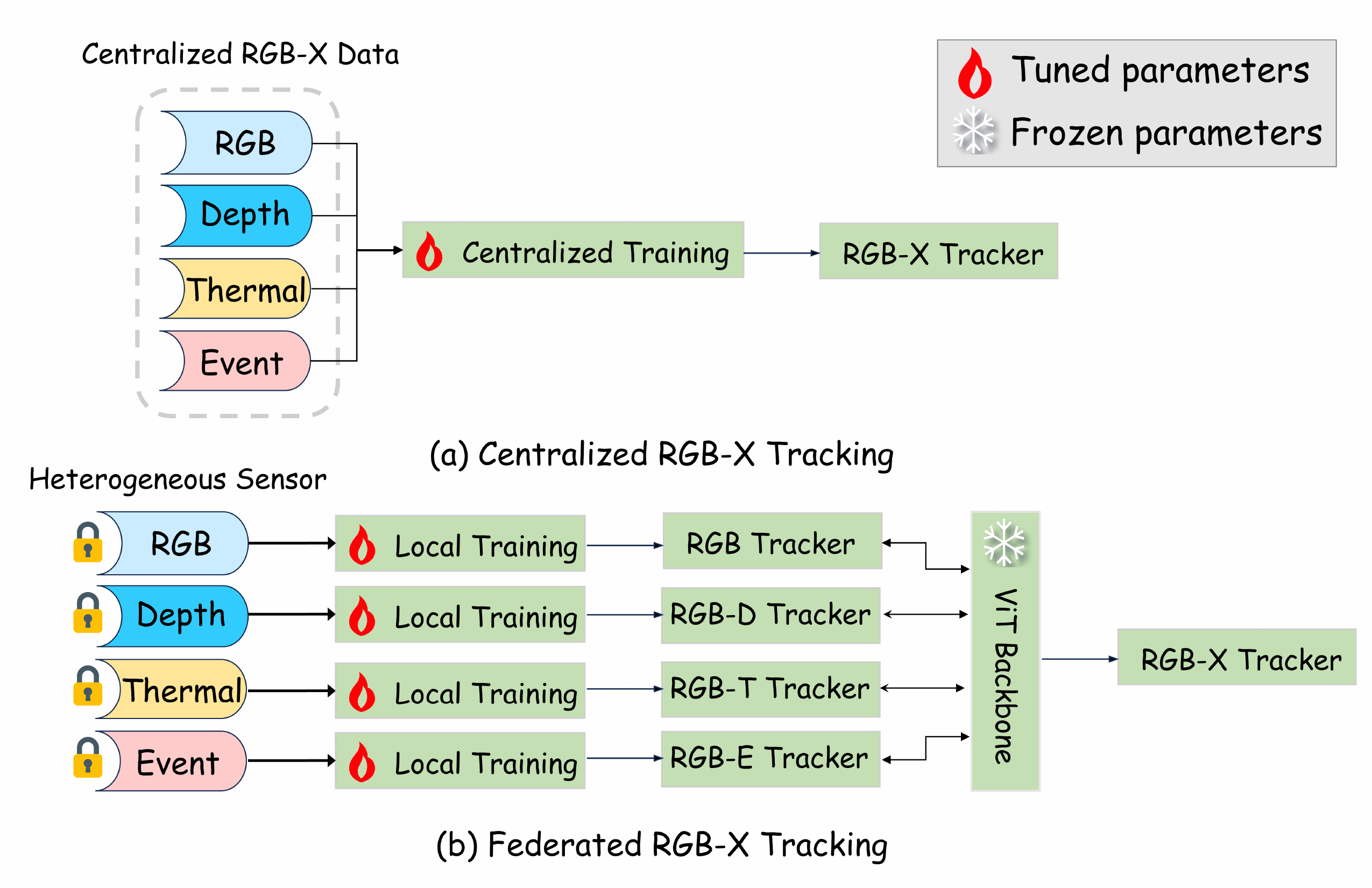}
  \caption{Motivation of federated RGB-X tracking. (a) Centralized RGB-X tracking pools aligned data from all sensors for one training run. (b) In the federated setting, clients with different sensors (RGB, depth, thermal, event) train locally, and only the tuned prompt parameters are aggregated on top of a frozen ViT backbone.}
  \label{fig:motivation}
\end{figure}

RGB-X trackers are commonly trained on centrally accessible paired videos~\cite{vipt,protrack}.
In a distributed setting, however, data may remain on devices with different sensor configurations.
Some clients have RGB-thermal, RGB-depth, or RGB-event observations, while others record RGB only.
Data-governance requirements and bandwidth or storage constraints may limit the transfer of these videos to a central training facility.
Federated learning~\cite{fedavg} provides an alternative in which clients keep videos local and exchange trainable parameters.
Figure~\ref{fig:motivation} illustrates the transition from centralized RGB-X training to this heterogeneous-client setting.
Keeping videos local avoids their direct transfer, although parameter exchange alone does not provide a formal privacy guarantee.

\textbf{Motivation.}
The central challenge is to determine which parameter groups clients should share and which clients should contribute to each group.
All clients observe RGB, making a common RGB pathway a natural sharing domain.
Auxiliary observations, in contrast, have different physical meanings across sensors, so a common parameter layout does not imply that every sensor-dependent update should be pooled globally.
Fusion and normalization parameters also interact with the auxiliary pathway, even when they are not explicitly labeled as auxiliary prompts.
Their sharing rules must therefore be considered alongside those for auxiliary-specific parameters.

A second distinction concerns RGB-only clients.
Their RGB observations remain useful for collaborative training.
However, fusion and normalization parameters can receive local updates even when auxiliary inputs are unavailable.
Assigning an RGB-only client to a modality bucket for parameter dispatch does not make its updates suitable for every parameter group in that bucket.
Sharing domains and contributor eligibility must therefore be defined explicitly and designed jointly.

Existing RGB-X methods primarily study local multimodal adaptation and fusion, whereas conventional and personalized federated methods address optimization or client-specific model components~\cite{fedprox,fedper,ditto}.
These approaches do not, by themselves, specify the sensor-aware sharing and participation rules needed for the different functional groups of an RGB-X tracker.
Separating the federation by modality avoids cross-sensor aggregation but also removes a shared RGB bank across modalities.
At the same time, exchanging the full tracker is unnecessary when adaptation can be confined to prompt-related parameters.
These observations motivate a framework that combines a common frozen tracker with group-specific federated aggregation.

We propose \methodname, a modality-aware federated prompt tuning framework built on a frozen pretrained tracker.
It organizes trainable parameters into RGB-shared, auxiliary-specific, fusion, and normalization groups.
The RGB-shared group is aggregated globally, while the other groups use thermal, event, and depth banks.
Selective participation completes the bucket design: RGB-only clients continue local training and contribute to the common RGB bank, but their fusion and normalization updates are excluded from modality-specific server averages.
Their auxiliary-specific parameters remain frozen and are already excluded from aggregation by the validity mask.
Bucketing and selective participation thus form a unified aggregation strategy.
Controlled experiments examine both design choices and show that extending the bucketing scope without adjusting contributor eligibility can reduce performance.

The main contributions of this paper are summarized as follows.
\begin{itemize}
\item We study federated RGB-X tracking with RGB-T, RGB-D, RGB-E, and RGB-only clients, explicitly distinguishing auxiliary-sensor availability from the originating dataset used for parameter dispatch.
\item We develop a prompt-only federated framework with a frozen tracking backbone and prediction head, restricting local optimization and round-wise communication to prompt-related parameters.
\item We jointly design modality-aware bucketing and selective participation: the former determines sharing domains, and the latter determines eligible contributors while retaining RGB-only participation in the shared RGB pathway.
\item We evaluate the framework on thermal, event, and depth benchmarks using matched training repetitions, controlled aggregation ablations, and final-parameter analyses, demonstrating improvements over FedAvg under the evaluated partition and clarifying the role of participation eligibility.
\end{itemize}

The remainder of this paper is organized as follows. Section~\ref{sec:related_work} reviews related work on RGB-X tracking, federated learning, and prompt tuning. Section~\ref{sec:method} presents the proposed \methodname framework. Section~\ref{sec:experiments} reports experimental settings, results, and ablation studies. Section~\ref{sec:discussion} discusses limitations and future directions, and Section~\ref{sec:conclusion} concludes this paper.

\section{Related Work}\label{sec:related_work}

We review four related areas: RGB-X visual tracking, federated learning for visual perception, multimodal federated learning, and parameter-efficient prompt tuning.
The comparison focuses on their treatment of sensor heterogeneity and parameter sharing in distributed RGB-X tracking.

\subsection{RGB-X Tracking}

RGB visual tracking localizes a target in successive frames given its initial state. Advances in Siamese networks and Transformer-based architectures have improved tracking accuracy and robustness through template-search matching, long-range dependency modeling, and feature interaction~\cite{siamfc,siamrpn,stark,ostrack}. However, RGB-only trackers are still vulnerable to challenging scenarios such as low illumination, heavy occlusion, motion blur, and complex background interference.

To overcome the limitations of RGB tracking, RGB-X tracking introduces auxiliary modalities, such as thermal infrared, depth, and event data, to provide complementary information. RGB-thermal tracking exploits illumination-insensitive thermal cues and is effective in low-light or nighttime scenes~\cite{rgbt_survey,lasher}. RGB-depth tracking uses geometric and spatial structure to distinguish targets from background distractors~\cite{depthtrack}. RGB-event tracking benefits from the high temporal resolution and motion sensitivity of event cameras, which can improve tracking under fast motion and illumination changes~\cite{visevent}. Existing RGB-X trackers mainly focus on cross-modal feature fusion, modality reliability estimation, and transformer-based multimodal interaction~\cite{vipt,protrack}. Although these methods have achieved promising performance, they are typically developed under a centralized training paradigm, assuming that aligned multimodal videos from different sensors can be collected in one place. Such an assumption is difficult to satisfy in real-world edge scenarios where raw videos are privacy-sensitive and clients have heterogeneous sensor configurations.

\subsection{Federated Visual Learning}

Federated learning enables multiple clients to train a model collaboratively without sharing their raw data. A typical approach is FedAvg, which aggregates local model updates on the server according to client data sizes or participation weights~\cite{fedavg}. To improve federated optimization under non-IID data, subsequent methods introduce proximal regularization, personalized model components, or client-specific adaptation strategies~\cite{fedprox,fedper,ditto}. These methods have been applied to various visual perception tasks, including image classification, object detection, and semantic segmentation.

Conventional model averaging typically uses a common parameterization across clients, but does not explicitly specify sensor-dependent sharing domains for RGB-X tracking.
Clients with RGB-thermal, RGB-depth, RGB-event, and RGB-only data can share tensor shapes while differing in the observations used to optimize those tensors.
Tracking also requires target-specific localization and template-search interaction rather than only image-level recognition.
These distinctions motivate an aggregation strategy that considers the functional roles of tracking parameters as well as sensor availability.

\subsection{Multimodal Federated Learning}

Multimodal federated learning studies collaborative training over distributed clients with multimodal data. Existing settings can be grouped into three categories according to the distribution of modalities across clients: all clients hold the same set of modalities, each client holds only one modality, or clients hold heterogeneous subsets of modalities~\cite{mmfl_survey}. To address modality heterogeneity, existing methods usually rely on representation alignment, knowledge distillation, modality completion, or personalized aggregation~\cite{creamfl,fedmultimodal}. These designs transfer knowledge across clients while reducing discrepancies caused by missing or heterogeneous modalities.
FedMSplit~\cite{Chen2022FedMSplit} divides client models into shareable blocks and uses a dynamic multi-view graph to model client correlations for block-wise aggregation.
FED-PRIME~\cite{Phung2025FedPrime} combines federated learning with multimodal prompt tuning, using inter-client prompts to transfer shared knowledge and intra-client prompts to model local missing-modality patterns.

Much of this work evaluates classification, recognition, or activity understanding, whereas RGB-X tracking uses auxiliary observations for target localization and template-search interaction.
Alignment and completion can facilitate knowledge transfer, but they do not by themselves determine whether RGB-shared, auxiliary-specific, fusion, and normalization parameters should use the same aggregation participants.
RGB-only clients make this distinction explicit: they have useful RGB data but no measurements of the corresponding auxiliary sensor.
Our focus is therefore a group-specific aggregation rule tailored to this tracking setting, rather than a replacement for multimodal alignment or completion.

\subsection{Parameter-Efficient Prompt Tuning}

Prompt tuning has emerged as an effective parameter-efficient adaptation technique. Instead of fine-tuning the entire model, prompt tuning learns a small set of prompt parameters while keeping most pretrained parameters frozen~\cite{vpt,prompt_tuning_survey}. This strategy has been explored in vision tasks to adapt large pretrained models with limited trainable parameters. For visual tracking, prompt-based adaptation can guide feature extraction, template-search interaction, and multimodal fusion, making it suitable for efficient RGB-X tracking~\cite{vipt}.

Prompt tuning also limits the parameter payload exchanged during federated training.
For personalized federated adaptation, pFedPG learns a server-side generator that produces client-specific visual prompts for frozen backbones~\cite{Yang2023pFedPG}.
In contrast to client-specific prompt generation, our focus is the sharing domain and contributor eligibility of each functional parameter group under heterogeneous sensor availability.
Our framework builds on prompt-based tracking and assigns aggregation rules to the functional groups of its trainable parameters.
This combines global RGB sharing with modality-specific adaptation and explicitly handles the eligibility of RGB-only clients.

\section{Method}\label{sec:method}

\methodname couples two aggregation decisions: where a parameter group is shared and which clients may contribute to that group.
We first define the local prompt tracker and its parameter groups.
Modality-aware bucketing then defines global and sensor-specific sharing domains, while selective participation determines the eligible contributors within each domain.
Together, bucketing and selective participation define a unified aggregation strategy.

\begin{figure*}[!t]
  \centering
  \includegraphics[width=.98\textwidth]{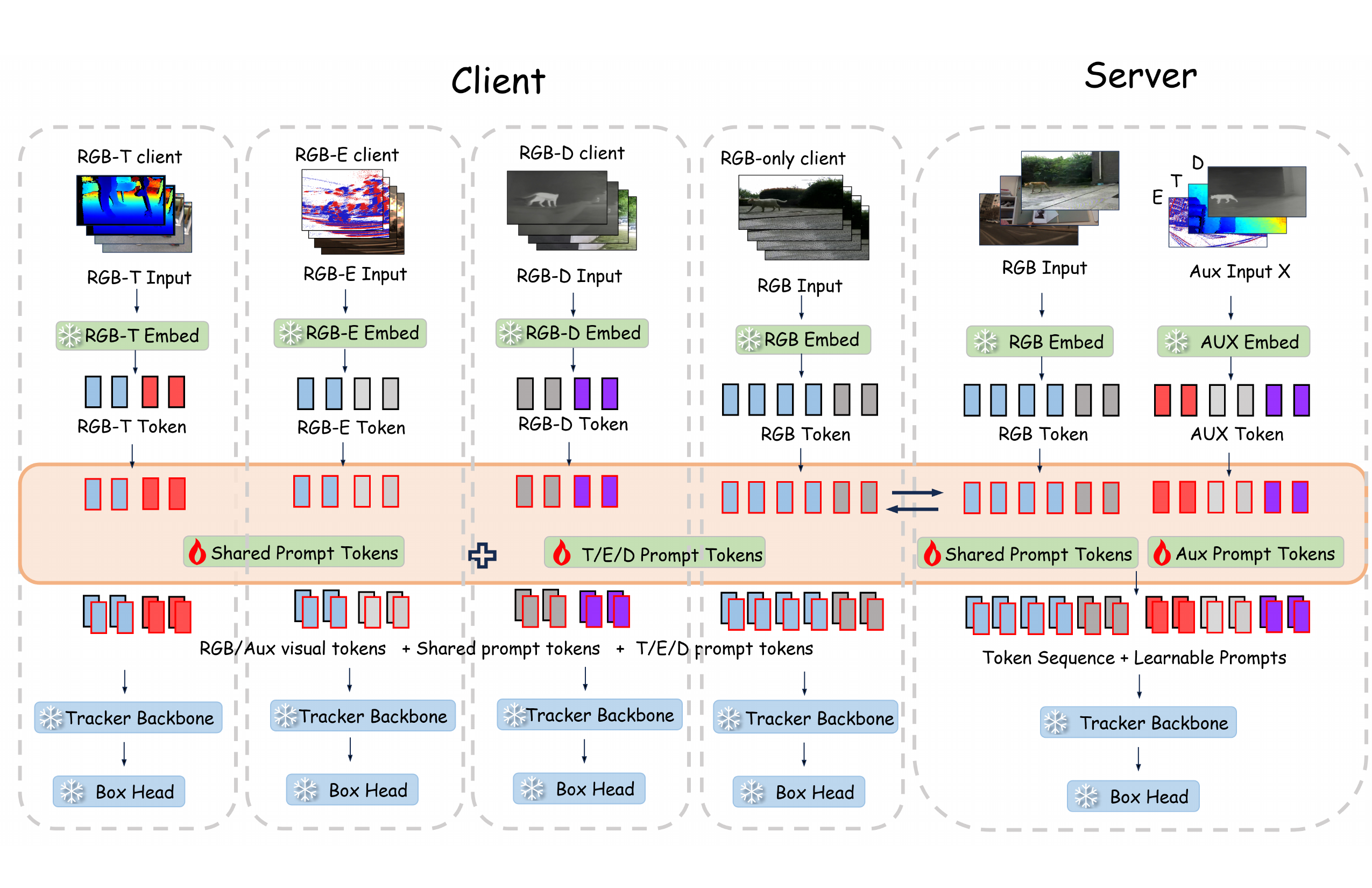}
  \caption{Overall framework of \methodname. Clients keep training videos local and adapt prompt-related parameters on a frozen tracking backbone and prediction head. RGB-shared parameters use one global bank, while auxiliary-specific, fusion, and normalization parameters use modality banks. Server aggregation combines this sharing structure with sensor-dependent contributor eligibility.}
  \label{fig:framework}
\end{figure*}

\subsection{Overall Framework}\label{subsec:framework}

\textbf{Clients and observations.}
Let $\mathcal{C}=\{1,\ldots,N\}$ be the client population and $\mathcal{S}^t\subseteq\mathcal{C}$ the clients participating in round $t$.
Client $i$ owns a local tracking dataset $\mathcal{D}_i$ with $n_i=|\mathcal{D}_i|$ assigned samples.
Its auxiliary-sensor type is $m_i\in\{T,D,E,\varnothing\}$, where $T$, $D$, and $E$ denote thermal, depth, and event sensors and $\varnothing$ denotes RGB only.
We write $v_i=\mathbb{I}(m_i\neq\varnothing)$ for client-level auxiliary availability.
This indicator describes the sensor configuration, not the presence of an auxiliary observation in every individual training sample.

A local sample contains a template $z$, a search observation $x$, and a target box $y$.
The template and search inputs each contain RGB and, when available, the corresponding auxiliary observation.
Unavailable auxiliary channels are zero-masked under the implemented input protocol.
The same tracker can therefore process the different client types without changing its parameter shapes.

\textbf{Functional parameter groups.}
Let $\Phi$ contain the frozen pretrained tracking backbone and prediction head.
The trainable prompt-related parameters are organized as
\begin{equation}
\Theta=\{\Theta^r,\Theta^a,\Theta^f,\Theta^n\},
\label{eq:prompt_groups}
\end{equation}
with group indices $\mathcal{G}=\{r,a,f,n\}$.
The RGB-shared group $r$ and auxiliary-specific group $a$ adapt the RGB and auxiliary pathways, respectively. The fusion group $f$ governs interactions between these pathways, while the normalization group $n$ contains the normalization parameters of the prompt modules.
The term ``prompt-related'' encompasses adapter and normalization tensors as well as prompt tokens.
This functional grouping determines the server's sharing and participation rules, while the local architecture remains identical across aggregation variants.

\textbf{Local adaptation.}
The client tracker predicts a target box as
\begin{equation}
\hat{y}=F_{\Phi}(z,x;\Theta_i).
\label{eq:tracker_forward}
\end{equation}
The local tracking objective is
\begin{equation}
\mathcal{L}_i(\Theta_i)=\frac{1}{n_i}\sum_{(z,x,y)\in\mathcal{D}_i}\ell\bigl(F_{\Phi}(z,x;\Theta_i),y\bigr),
\label{eq:local_objective}
\end{equation}
where $\ell$ is the base tracker's original loss, which combines box regression with target-response supervision.
Mini-batches are sampled from the client's assigned sequences using the same preprocessing protocol across the core comparisons.

Local trainability and server eligibility are separate concepts.
For the configuration evaluated here, define the local trainability indicator as
\begin{equation}
u_i^g=\begin{cases}
v_i, & g=a,\\
1, & g\in\{r,f,n\}.
\end{cases}
\label{eq:local_trainability}
\end{equation}
Thus, an RGB-only client freezes its auxiliary-specific group but can locally optimize RGB-shared, fusion, and normalization parameters with masked auxiliary inputs.
A parameter group may therefore be trainable locally without being eligible for server aggregation.

At the start of round $t$, the client receives the server prompt view $\Theta_i^{t,0}$.
Its final local parameters are denoted by
\begin{equation}
\widehat{\Theta}_i^t
=\operatorname{LocalAdamW}_i\bigl(\Theta_i^{t,0};\Phi,E\bigr),
\label{eq:local_update}
\end{equation}
where $E$ is the local mini-batch budget and only groups with $u_i^g=1$ are optimized.
The operator denotes budgeted local AdamW training, including the gradient accumulation specified in Section~\ref{subsec:exp_setup}.
The client returns only its prompt tensors and validity metadata, keeping raw videos and frozen model parameters local.

\subsection{Modality-Aware Aggregation}\label{subsec:bucket}

\textbf{From one sharing domain to sensor-specific domains.}
A global prompt baseline aggregates valid updates from all participating clients.
Let $q_i^{g,t}$ indicate whether client $i$ supplies a valid update for group $g$ in round $t$, and let $\mathcal{V}_g^t=\{i\in\mathcal{S}^t:q_i^{g,t}=1\}$.
For a nonempty valid set, global aggregation is
\begin{equation}
\Theta_{\mathrm{global}}^{g,t+1}
=\sum_{i\in\mathcal{V}_g^t}
\frac{n_i}{\sum_{j\in\mathcal{V}_g^t}n_j}
\widehat{\Theta}_i^{g,t}.
\label{eq:fedavg}
\end{equation}
Validity is checked before averaging.
For example, an RGB-only client's frozen auxiliary-specific tensors do not become valid contributions merely because they have the expected shape.
In the implementation, these checks are applied to the uploaded tensors using parameter masks.

Global aggregation assigns a single sharing domain to each parameter group.
This choice is restrictive for RGB-X tracking because thermal appearance, depth geometry, and brightness-change events may require different processing and interactions with RGB.
Bucketing therefore assigns sensor-dependent groups to distinct domains rather than assuming that all such tensors should share one bank.
This separates sensor-specific adaptation from common RGB sharing without changing the local model.

\textbf{Bucket assignment and dispatch.}
The bucket label is
\begin{equation}
b_i=\begin{cases}
m_i, & m_i\neq\varnothing,\\
\rho(i), & m_i=\varnothing,
\end{cases}
\label{eq:bucket_label}
\end{equation}
where $\rho(i)\in\{T,D,E\}$ identifies the auxiliary modality of the client's originating dataset.
This metadata specifies the parameter configuration dispatched to an RGB-only client without implying that the corresponding sensor is available.
For parameter group $g$, define
\begin{equation}
\Omega_g(i)=\begin{cases}
\star, & g=r,\\
b_i, & g\in\{a,f,n\},
\end{cases}
\label{eq:aggregation_domain}
\end{equation}
where $\star$ denotes the global RGB bank.
The server therefore maintains one global RGB-shared bank and three modality banks for each remaining parameter group.
The initialization dispatched to a client is
\begin{equation}
\Theta_i^{g,t,0}=\begin{cases}
\Theta_{\mathrm{init}}^a, & g=a,\ v_i=0,\\
\Theta_{\Omega_g(i)}^{g,t}, & \text{otherwise}.
\end{cases}
\label{eq:prompt_dispatch}
\end{equation}
Here, $\Theta_{\mathrm{init}}^a$ is the stored initial auxiliary-specific state.
It is dispatched and held fixed for RGB-only clients, while their fusion and normalization parameters use the banks associated with their originating datasets.
Dispatch determines the client's initial parameter configuration, while aggregation eligibility determines whether its returned tensors contribute to the corresponding bank.

\textbf{Group-specific weighted aggregation.}
Let $a_i^g$ be the server participation gate defined in Section~\ref{subsec:selective}.
For domain $d$ and group $g$, the effective contributor set is
\begin{equation}
\mathcal{C}_d^{g,t}=\{i\in\mathcal{S}^t\mid\Omega_g(i)=d,\ q_i^{g,t}=1,\ a_i^g=1\}.
\label{eq:contributor_set}
\end{equation}
A client belongs to this set if it participates in the current round, matches the sharing domain, and supplies a valid update that passes the participation gate.
If it is nonempty, the server computes
\begin{equation}
\Theta_d^{g,t+1}
=\sum_{i\in\mathcal{C}_d^{g,t}}
\frac{n_i}{\sum_{j\in\mathcal{C}_d^{g,t}}n_j}
\widehat{\Theta}_i^{g,t}.
\label{eq:bucket_aggregation}
\end{equation}
The denominator is recomputed for the eligible group-domain pair.
Excluded updates are not represented by zero tensors, and they do not retain weight in the normalization.
If no eligible update is available, the server instead preserves the previous bank:
\begin{equation}
\Theta_d^{g,t+1}=\Theta_d^{g,t},
\qquad\mathcal{C}_d^{g,t}=\emptyset.
\label{eq:keep_previous}
\end{equation}
This handles absent contributors without overwriting a modality bank with unrelated updates or a zero-valued replacement.

\begin{figure}[pos=t]
  \centering
  \includegraphics[width=\columnwidth]{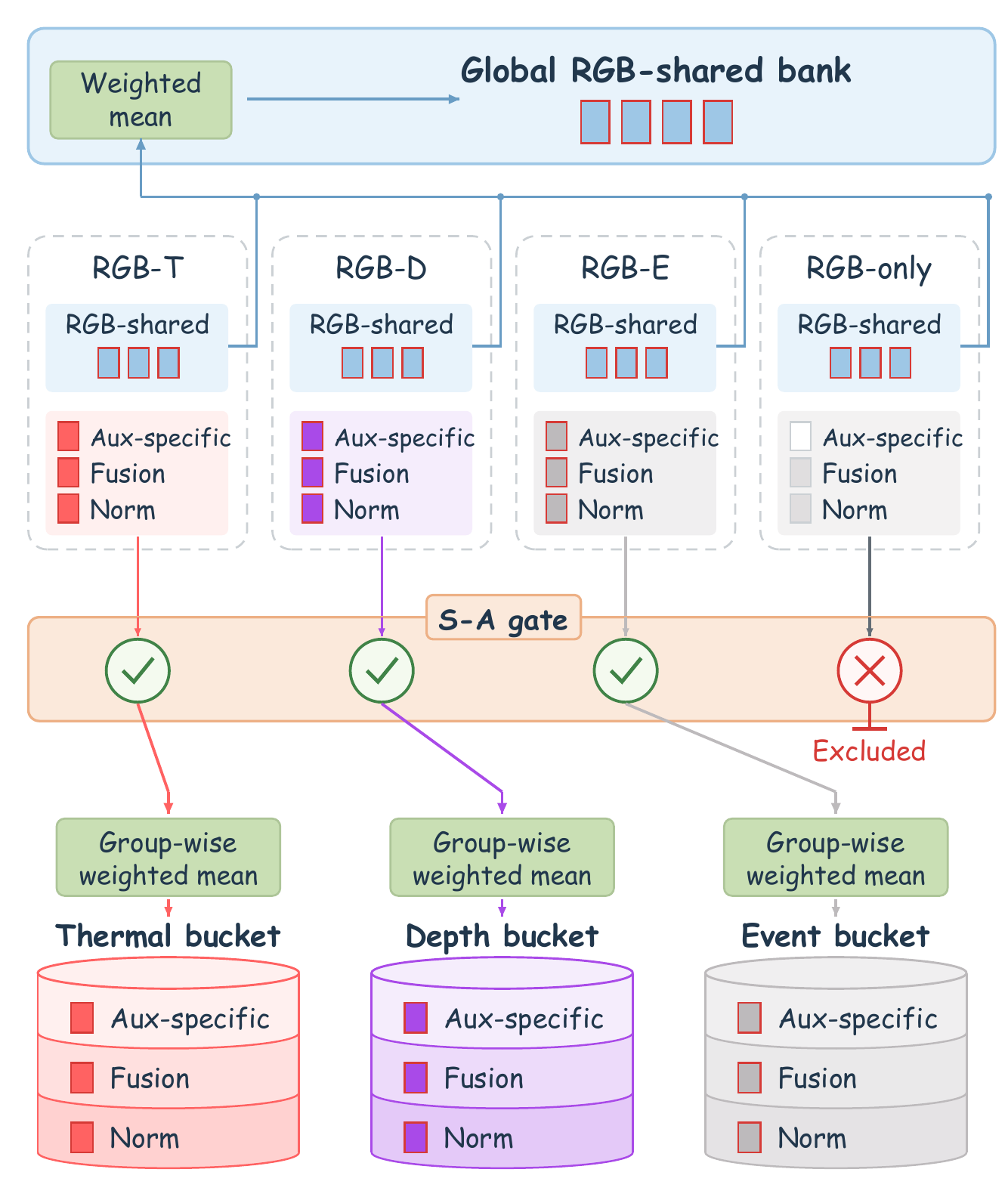}
  \caption{Unified aggregation design. RGB-shared parameters use the global bank, while auxiliary-specific, fusion, and normalization parameters use thermal, depth, and event banks. The participation gate retains valid RGB updates from RGB-only clients but excludes their modality-related contributions from server bucket averages. Dispatch and local trainability are distinct from server eligibility.}
  \label{fig:aggregation_details}
\end{figure}

\subsection{Selective Participation}\label{subsec:selective}

Bucketing identifies a sharing domain but does not establish the suitability of every contributor assigned to that domain.
An RGB-only client can locally change fusion and normalization parameters using its masked auxiliary inputs, yet those changes are not learned from observed RGB-X interaction.
At the same time, its RGB observations remain relevant to the shared RGB pathway.
Selective participation therefore acts on parameter groups rather than removing the client from collaborative training.

The server participation gate is
\begin{equation}
a_i^g=\begin{cases}
1, & g=r,\\
v_i, & g\in\{a,f,n\}.
\end{cases}
\label{eq:selective_rule}
\end{equation}
Together with Eq.~\eqref{eq:contributor_set}, this retains every valid RGB contribution while restricting modality-related server averages to equipped clients in the matching bucket.
The validity mask $q_i^{g,t}$ still applies: sensor availability does not override an invalid uploaded tensor.
Occasional sample-level auxiliary masking does not, by itself, change the client's sensor label.

For auxiliary-specific parameters, invalid RGB-only contributions are already rejected by the validity mask.
The additional effect of S-A in the controlled FNA comparison is therefore the exclusion of RGB-only fusion and normalization updates.
FNA uses the same bank assignment and dispatch rule but sets $a_i^g=1$ for all groups, leaving those locally valid updates in the averages.
The comparison therefore isolates server aggregation eligibility while holding sharing domains and the local training protocol fixed.

An RGB-only client still receives its dataset-associated bank, trains its permitted local groups, and supplies its RGB-shared update.
Its fusion and normalization tensors are filtered from the corresponding server averages, not necessarily omitted from the upload.
Thus, S-A is not an additional communication-compression mechanism.
Prompt-only exchange reduces the tensor payload, while the aggregation strategy determines how the exchanged tensors are combined.
Algorithm~\ref{alg:fedmaptrack} summarizes the complete procedure. Section~\ref{sec:experiments} evaluates tracking performance and quantifies the communication payload.

\begingroup
\algtext*{EndFor}
\algtext*{EndIf}
\begin{algorithm}[t]
\caption{Unified federated prompt training with bucketing and S-A.}
\label{alg:fedmaptrack}
\footnotesize
\begin{algorithmic}[1]
\Require Clients $\mathcal{C}$ and local datasets $\mathcal{D}_i$;
\Require Sensor labels $m_i$ and dataset labels $\rho(i)$;
\Require Frozen tracker $\Phi$, rounds $R$, mini-batch budget $E$.
\Ensure Global RGB bank and modality-related banks.
\State Initialize all banks from the common prompt state.
\State Store the fixed auxiliary state $\Theta_{\mathrm{init}}^a$.
\State Define $\mathcal{B}_r=\{\star\}$ and $\mathcal{B}_g=\{T,D,E\}$ for $g\neq r$.
\State Set client buckets by Eq.~\eqref{eq:bucket_label}.
\For{$t=0$ to $R-1$}
  \State Select participating clients $\mathcal{S}^t$.
  \ForAll{$i\in\mathcal{S}^t$ in parallel}
    \State Download the view in Eq.~\eqref{eq:prompt_dispatch}.
    \State Set trainability by Eq.~\eqref{eq:local_trainability}.
    \State Run local AdamW in Eq.~\eqref{eq:local_update}.
    \State Return prompt tensors and validity metadata.
  \EndFor
  \ForAll{$g\in\mathcal{G},\ d\in\mathcal{B}_g$}
    \State Apply the gate in Eq.~\eqref{eq:selective_rule}.
    \State Build $\mathcal{C}_d^{g,t}$ by Eq.~\eqref{eq:contributor_set}.
    \If{$\mathcal{C}_d^{g,t}\neq\emptyset$}
      \State Aggregate by Eq.~\eqref{eq:bucket_aggregation}.
    \Else
      \State Keep the bank by Eq.~\eqref{eq:keep_previous}.
    \EndIf
  \EndFor
\EndFor
\State \Return Global RGB bank and the modality banks.
\end{algorithmic}
\end{algorithm}
\endgroup

\section{Experiments}
\label{sec:experiments}

\begingroup
\setlength{\emergencystretch}{2em}

\subsection{Experimental Setup}
\label{subsec:exp_setup}

\textbf{Implementation and training configuration.}
We implement \methodname\ on the ViPT tracking architecture~\cite{vipt}.
The pretrained tracking backbone and prediction head are frozen throughout federated optimization, while only prompt-related parameters are updated locally.
Each sensor-equipped local model contains 839,628 trainable parameters, organized into RGB-shared, auxiliary-specific, fusion, and normalization groups with 73,836, 664,416, 82,944, and 18,432 parameters, respectively.
All core aggregation variants use this parameterization, so their performance differences cannot be attributed to increased local model capacity.
The training configurations specify NVIDIA RTX 4090 GPUs, while all benchmark evaluations use the same machine equipped with an NVIDIA RTX 4080 SUPER GPU.

The core experiments use 120 communication rounds, with all twelve clients selected in each round.
Each selected client processes 50 local mini-batches per round using AdamW, a learning rate of $4\times10^{-4}$, and weight decay of $10^{-4}$.
The configured mini-batch size is 32, with gradient accumulation over two mini-batches, giving an effective optimization batch size of 64 and 25 optimizer updates per client per round.
We distinguish local mini-batches from optimizer updates because the implementation's local-step budget counts the former.
Optimizer moments are reset at the start of each local training round.
Template and search crops have spatial resolutions of $128\times128$ and $256\times256$, respectively.
All core variants use the same tracking objective, preprocessing, local optimization budget, and parameter initialization for matched training repetitions.
Progressive prompt-group warm-up is not used in these comparisons, so the ablation isolates aggregation rather than an additional training schedule.

\textbf{Datasets and client partition.}
We evaluate RGB-thermal, RGB-event, and RGB-depth tracking on LasHeR~\cite{lasher}, VisEvent~\cite{visevent}, and DepthTrack~\cite{depthtrack}, respectively.
Training sequences are assigned to clients using one fixed sequence-level Dirichlet partition with concentration parameter $\alpha=0.5$.
The partition is constructed within each benchmark and remains unchanged across the compared methods.
Each benchmark supplies four clients: three sensor-equipped RGB-X clients and one RGB-only client.
The resulting federation contains nine sensor-equipped clients and three RGB-only clients.
Training sequences remain local to their assigned clients, while evaluation uses the corresponding benchmark test split.
Client construction does not use manually assigned scene categories.

An RGB-only client retains the RGB observations of its assigned sequences but has no usable auxiliary sensor input.
Its auxiliary-specific parameter group is frozen and excluded from valid auxiliary aggregation.
For sensor-equipped clients, auxiliary observations are additionally masked with probability 0.1 during local training, using the same masking protocol in the core comparisons.
This sample-level augmentation is distinct from client-level sensor availability: a sensor-equipped client remains eligible under S-A despite occasional masking, whereas an RGB-only client has no auxiliary observations throughout training.
The assigned training-sample counts are 459,122 for LasHeR, 212,062 for VisEvent, and 209,913 for DepthTrack, giving 881,097 samples in total.
The dataset-specific allocation to RGB-only clients is analyzed in Section~\ref{subsec:rgb_only_analysis}.

\begingroup
\setlength{\emergencystretch}{2em}
\textbf{Evaluation protocol.}
We report success-curve AUC for LasHeR, success rate (SR) for VisEvent, and the official F-score for DepthTrack.
The LasHeR evaluator averages the success curve over overlap thresholds from 0 to 1 in increments of 0.05.
In the VisEvent evaluator, SR is the mean of the success curve on the same threshold grid, rather than success at a single overlap threshold.
DepthTrack is evaluated using its benchmark analysis protocol.
All three metrics are expressed as percentages, with higher values indicating better tracking performance.
The benchmarks differ in their metric definitions, so we report them separately and do not calculate a combined score.

For every run, we evaluate the checkpoint saved at communication round 120.
Checkpoints are not selected according to test performance.
Each main-table method and core ablation includes three matched training runs with seeds 42, 0, and 123.
We report the mean and sample standard deviation across these runs, using $n-1$ as the variance denominator.
For Local, FedPer, Ditto, FedMSplit-P, and FED-PRIME, each sensor-equipped client has its own evaluated model.
Within each repetition, we average the benchmark scores of its three sensor-equipped client models. We then compute the reported mean and sample standard deviation across the three repetitions.
For the remaining main-table methods, each benchmark is evaluated using its corresponding server-side prompt configuration.
This distinguishes an average of client-model scores from the score of a single aggregated model and avoids treating the nine client evaluations as nine independent training repetitions.
The same client partition is used in every repetition.
Consequently, the reported standard deviations describe training variability under the evaluated partition and do not measure variability across independently generated client partitions.
Differences discussed below are calculated from the displayed mean scores and expressed in percentage points.
\par
\endgroup

For the supplementary LasHeR curves, all methods use the same 245 test sequences.
Success curves use the IoU threshold grid specified above, and precision curves use location-error thresholds from 0 to 50 pixels in one-pixel increments.
We first average the curves over the test sequences within each repetition. The displayed curves and shaded bands then represent the mean and sample standard deviation across the three repetitions.
Precision at 20 pixels provides a complementary center-localization metric and does not replace the primary success AUC comparison.

\textbf{Compared methods.}
The main comparison includes \textbf{Local}, \textbf{FedAvg}~\cite{fedavg}, \textbf{FedProx}~\cite{fedprox}, \textbf{FedPer}~\cite{fedper}, \textbf{Ditto}~\cite{ditto}, \textbf{FedMSplit-P}~\cite{Chen2022FedMSplit,Phung2025FedPrime}, \textbf{FED-PRIME}~\cite{Phung2025FedPrime}, \textbf{Independent FL}, and the complete \textbf{\methodname}.
Local provides a non-collaborative reference using separately trained client models.
FedAvg performs sample-weighted global averaging over valid prompt updates without maintaining modality-specific parameter banks.
FedProx adds a proximal penalty with coefficient $\mu=0.01$ to constrain local deviation from the server parameters.
FedPer provides a shared-and-personalized model reference, and Ditto combines a federated model with client-specific regularized models using $\lambda=0.1$.
FedMSplit-P and FED-PRIME are adapted to the ViPT tracker following the prompt-tuning implementations in~\cite{Phung2025FedPrime}.
FedMSplit-P maintains a model for each client and replaces uniform server averaging with a similarity-weighted combination of client models for each parameter group, using a proximal coefficient of 0.01.
FED-PRIME maintains intra-client and inter-client prompt pools of 20 tokens each and prepends the top five tokens from each pool to the Transformer input.
The intra-client pool remains local, while inter-client prompts are aligned by Hungarian matching and averaged across clients. The remaining parameter groups follow FedAvg aggregation.
Hyperparameters follow the original implementation defaults and are not tuned for individual benchmarks.
These are adapted tracking baselines rather than exact reproductions of the original methods on their classification benchmarks.
These coefficients are fixed across the three benchmarks and training repetitions in the reported comparison.
The client-model evaluation of Local, FedPer, Ditto, FedMSplit-P, and FED-PRIME follows the reporting rule stated above.
Independent FL constructs a separate federation for each benchmark using its three sensor-equipped clients and one RGB-only client, with no prompt parameters shared across federations.
\methodname\ maintains one global RGB-shared bank and modality-specific auxiliary, fusion, and normalization banks, with selective participation applied to the modality-interactive groups.
Both the sharing domains and aggregation eligibility are therefore determined at the parameter-group level.
Auxiliary-only bucketing and extended bucketing without S-A are controlled variants of this design and appear in the ablation rather than as competing final methods in the main table.

The experiments are organized around four research questions:
\begin{itemize}
  \item \textbf{RQ1:} How does the complete framework compare with the available federated tracking baselines across auxiliary modalities?
  \item \textbf{RQ2:} How do the bucketing scope and selective participation affect tracking performance?
  \item \textbf{RQ3:} How does the participation design relate to RGB-only data allocation and the final structure of modality-related parameters?
  \item \textbf{RQ4:} What local optimization and localization behavior accompanies the quantitative results?
\end{itemize}
We additionally account for the prompt communication payload, distinguishing it from the effect of the aggregation design.

\subsection{Main Results}
\label{subsec:main_results}

\begin{table*}[!t]
  \centering
  \caption{Main results on RGB-X tracking benchmarks at communication round 120. Scores are mean $\pm$ sample standard deviation across three matched training repetitions under one fixed client partition. For Local, FedPer, Ditto, FedMSplit-P, and FED-PRIME, each repetition first averages the scores of the benchmark's three sensor-equipped client models. Bold indicates the highest mean in each metric.}
  \label{tab:main_results}
  \begingroup
  \small
  \renewcommand{\arraystretch}{1.18}
  \setlength{\tabcolsep}{5pt}
  \begin{tabular*}{\textwidth}{@{\extracolsep{\fill}}llccc@{}}
    \toprule
    Method & Category & \makecell{LasHeR (RGB-T)\\AUC (\%)} & \makecell{VisEvent (RGB-E)\\SR (\%)} & \makecell{DepthTrack (RGB-D)\\F-score (\%)} \\
    \midrule
    Local & Local training & $47.34 \pm 0.31$ & $52.27 \pm 0.25$ & $49.91 \pm 1.46$ \\
    FedAvg & Standard FL & $46.55 \pm 0.99$ & $52.93 \pm 0.59$ & $53.85 \pm 0.73$ \\
    FedProx & Standard FL & $46.78 \pm 0.67$ & $53.11 \pm 0.66$ & $54.06 \pm 0.64$ \\
    FedPer & Personalized FL & $47.35 \pm 0.54$ & $54.47 \pm 0.36$ & $50.81 \pm 0.68$ \\
    Ditto & Personalized FL & $46.46 \pm 0.86$ & $54.14 \pm 0.42$ & $53.90 \pm 0.58$ \\
    FedMSplit-P & Multimodal FL & $46.43 \pm 0.33$ & $54.85 \pm 0.35$ & $54.24 \pm 0.89$ \\
    FED-PRIME & Multimodal FL & $46.78 \pm 0.45$ & $52.09 \pm 0.26$ & $54.32 \pm 0.17$ \\
    Independent FL & Separate FL & $48.25 \pm 0.65$ & $\mathbf{56.45 \pm 0.83}$ & $55.28 \pm 1.35$ \\
    \midrule
    \methodname & Ours & $\mathbf{48.90 \pm 0.63}$ & $56.38 \pm 0.41$ & $\mathbf{55.96 \pm 0.50}$ \\
    \bottomrule
  \end{tabular*}
  \endgroup
\end{table*}

\textbf{Comparison with local and modality-agnostic training.}
Table~\ref{tab:main_results} addresses RQ1.
\methodname\ achieves mean scores of 48.90 AUC on LasHeR, 56.38 SR on VisEvent, and 55.96 F-score on DepthTrack.
Compared with FedAvg, the corresponding gains are 2.35, 3.45, and 2.11 percentage points.
The improvement over FedAvg occurs in each of the three matched training repetitions on all three benchmarks.
Because the tracking architecture, local parameter capacity, training budget, client partition, and checkpoint rule are held fixed, these results provide evidence for the proposed aggregation design under the evaluated heterogeneous-client setting.
They indicate that globally pooling valid prompt updates is less effective here than assigning appropriate sharing domains and participants to the different parameter groups.

The complete method also exceeds Local by 1.56 thermal AUC points, 4.11 event SR points, and 6.05 depth F-score points.
However, collaboration alone does not improve every benchmark: FedAvg has lower thermal AUC than Local despite higher event and depth scores.
FedProx improves the FedAvg means by 0.23 thermal AUC points, 0.18 event SR points, and 0.21 depth F-score points, but remains below \methodname\ by 2.12, 3.27, and 1.90 percentage points, respectively.
With the reported coefficient $\mu=0.01$, proximal regularization does not match the complete aggregation design.
This comparison concerns the evaluated configuration and does not establish the relative performance of all possible proximal coefficients.

\textbf{Comparison with personalized federation.}
\methodname\ obtains higher mean scores than both FedPer and Ditto on all three benchmarks.
The thermal, event, and depth gains are 1.55, 1.91, and 5.15 percentage points over FedPer and 2.44, 2.24, and 2.06 percentage points over Ditto, respectively.
Personalization does not produce a uniform advantage over FedAvg: FedPer improves thermal and event performance but has lower depth performance, while Ditto has lower thermal performance and higher event and depth means.
Under the evaluated settings, generic personalization is not sufficient to reproduce the mean performance of the proposed sensor-aware aggregation strategy.
FedPer and Ditto are evaluated by averaging client-model scores, as described above. Their model outputs and deployment configurations therefore differ from those of methods evaluated using server-side banks.

\textbf{Comparison with multimodal federated baselines.}
\methodname\ outperforms the adapted FedMSplit-P and FED-PRIME baselines on all three benchmarks in each matched training repetition.
Compared with FedMSplit-P, the gains are 2.47 thermal AUC points, 1.53 event SR points, and 1.72 depth F-score points. Compared with FED-PRIME, the corresponding gains are 2.12, 4.29, and 1.64 points.
These results support group-specific aggregation in the evaluated tracking setting. Because both baselines are adapted implementations, the comparison does not characterize their performance on their original tasks.

\textbf{Comparison with modality-separated federation.}
Relative to Independent FL, \methodname\ increases mean thermal AUC from 48.25 to 48.90 and mean depth F-score from 55.28 to 55.96, corresponding to gains of 0.65 and 0.68 percentage points.
The mean event SR is slightly lower, at 56.38 compared with 56.45, a difference of 0.07 percentage points.
Thus, among all nine main-table methods, the complete method obtains the highest mean thermal and depth scores, while Independent FL obtains the highest event mean.
The comparison shows that retaining a common RGB-shared bank is compatible with competitive tracking across the three modalities, while modality-related banks preserve their distinct aggregation domains.
Since the two methods also differ in their fusion and normalization participation rules, this comparison alone does not isolate the contribution of cross-modal RGB sharing.

Overall, the complete method has higher mean scores than Local, FedAvg, FedProx, FedPer, Ditto, FedMSplit-P, and FED-PRIME on all three benchmarks, but does not improve the event mean over Independent FL.
The result therefore supports improved mean tracking performance over the evaluated modality-agnostic and personalized references, with benchmark-dependent differences relative to separate federations.
The sample standard deviations describe training variability under the fixed partition. Neither the reported gains nor the small event difference establishes statistical significance or equivalence.

\begin{figure*}[!t]
  \centering
  \includegraphics[width=\textwidth]{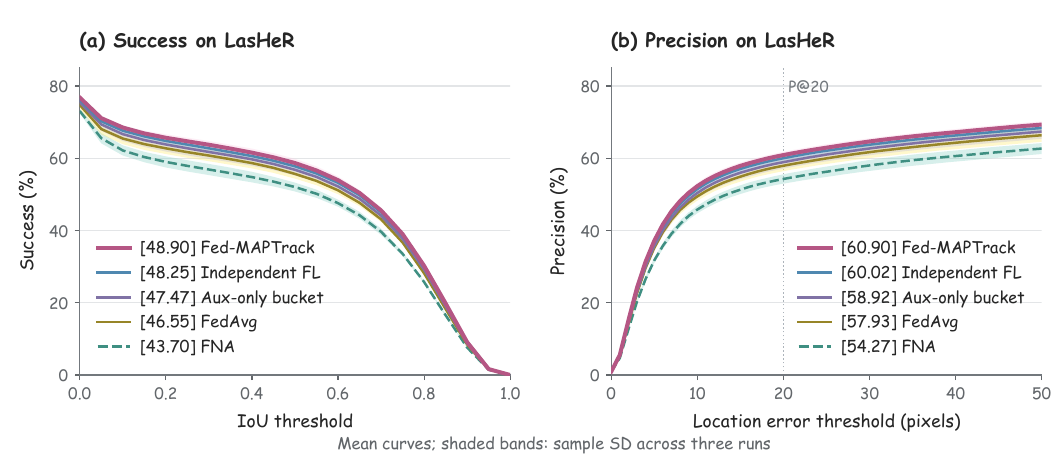}
  \caption{Success and precision curves on LasHeR at communication round 120 using the same 245 test sequences. Curves show means across three training repetitions, and shaded bands indicate sample standard deviations. Legend values are success AUC in (a) and precision at a 20-pixel threshold in (b), expressed as percentages. Auxiliary-only bucketing and FNA are included as controlled aggregation variants.}
  \label{fig:lasher_tracking_curves}
\end{figure*}

\textbf{Success and precision across thresholds.}
Figure~\ref{fig:lasher_tracking_curves} complements the scalar LasHeR results with the complete measured curves.
The complete method achieves the highest mean AUC and precision at 20 pixels among the five displayed configurations.
Its precision is 60.90\%, compared with 57.93\% for FedAvg, 54.27\% for FNA, and 60.02\% for Independent FL.
The precision ordering is consistent with the AUC ordering and provides a complementary view of center localization.
The curves reveal how the aggregation variants differ across the full threshold ranges.
These are additional views of the same test evaluations, not independent experiments or evidence of uniform dominance at every threshold.

\subsection{Ablation Study}
\label{subsec:ablation}

To address RQ2, we vary the aggregation configuration while retaining the same twelve clients, local training budget, parameter initialization protocol, and evaluation at communication round 120.
Table~\ref{tab:ablation} compares four configurations.
\textbf{FedAvg} provides the reference with global prompt aggregation.
\textbf{Auxiliary-only bucketing} aggregates auxiliary-specific parameters within thermal, event, and depth banks, while RGB-shared, fusion, and normalization parameters remain globally aggregated.
\textbf{Extended bucketing without S-A (FNA)} also assigns fusion and normalization parameters to modality-specific banks, but allows RGB-only clients to contribute to these two banks according to their originating dataset.
\textbf{\methodname} uses the same banks as FNA and additionally excludes RGB-only fusion and normalization updates from the server's bucket averages.
RGB-only clients continue to contribute to global RGB-shared aggregation.
Auxiliary-specific aggregation excludes invalid RGB-only updates in all auxiliary-bucket configurations, including auxiliary-only bucketing and FNA.

\begin{table*}[!t]
  \centering
  \caption{Ablation of the aggregation design. All scores are mean $\pm$ sample standard deviation over three matched training repetitions. Checkmarks indicate enabled components. RGB-shared parameters remain globally aggregated in all four configurations. FNA and \methodname\ have identical bucketing scopes and differ only in RGB-only eligibility for fusion and normalization aggregation. Bold indicates the highest mean in each metric.}
  \label{tab:ablation}
  \begingroup
  \small
  \renewcommand{\arraystretch}{1.2}
  \setlength{\tabcolsep}{4pt}
  \begin{tabular*}{\textwidth}{@{\extracolsep{\fill}}lcccccc@{}}
    \toprule
    \multirow{2}{*}{Variant} & \multicolumn{3}{c}{Aggregation component} & \multicolumn{3}{c}{Tracking performance} \\
    \cmidrule(lr){2-4}\cmidrule(lr){5-7}
    & \makecell{Auxiliary\\bucket} & \makecell{Fusion/Norm\\bucket} & \makecell{Selective\\participation}
    & \makecell{RGB-T\\AUC (\%)} & \makecell{RGB-E\\SR (\%)} & \makecell{RGB-D\\F-score (\%)} \\
    \midrule
    FedAvg & $\times$ & $\times$ & $\times$ & $46.55 \pm 0.99$ & $52.93 \pm 0.59$ & $53.85 \pm 0.73$ \\
    \makecell[l]{Auxiliary-only\\bucketing} & $\checkmark$ & $\times$ & $\times$ & $47.47 \pm 1.24$ & $\mathbf{56.46 \pm 0.37}$ & $\mathbf{56.02 \pm 0.32}$ \\
    \makecell[l]{Extended bucketing\\without S-A (FNA)} & $\checkmark$ & $\checkmark$ & $\times$ & $43.70 \pm 0.95$ & $54.21 \pm 0.27$ & $55.31 \pm 0.57$ \\
    \methodname & $\checkmark$ & $\checkmark$ & $\checkmark$ & $\mathbf{48.90 \pm 0.63}$ & $56.38 \pm 0.41$ & $55.96 \pm 0.50$ \\
    \bottomrule
  \end{tabular*}
  \endgroup
\end{table*}

\textbf{Separating auxiliary-specific parameters.}
Auxiliary-only bucketing improves the mean scores over FedAvg by 0.92 thermal AUC points, 3.53 event SR points, and 2.17 depth F-score points.
The improvement is achieved without isolating RGB-shared, fusion, or normalization parameters, showing that protecting the aggregation domain of auxiliary-specific parameters is useful under the evaluated partition.
This is consistent with the design motivation that thermal, event, and depth observations provide different auxiliary information and should not automatically share one auxiliary prompt bank.
This experiment evaluates the effect of auxiliary-specific bucketing on tracking performance, without directly measuring gradient conflict between modalities.

\textbf{Extending the bucketing scope.}
Adding modality-specific fusion and normalization banks without modifying participation eligibility reduces mean performance relative to auxiliary-only bucketing.
The thermal, event, and depth scores decrease from 47.47, 56.46, and 56.02 to 43.70, 54.21, and 55.31, respectively.
Thus, increasing the number of bucketed parameter groups is not sufficient to improve tracking.
Unlike auxiliary-specific groups, whose invalid RGB-only updates are already masked, fusion and normalization groups remain locally trainable for RGB-only clients.
Under FNA, those updates enter the corresponding dataset bucket despite being learned without auxiliary observations.
This distinction motivates treating the sharing domain and contributor eligibility as coupled design choices.

\textbf{Adding selective participation.}
With the bucketing scope fixed, \methodname\ improves FNA by 5.20 thermal AUC points, 2.17 event SR points, and 0.65 depth F-score points.
Figure~\ref{fig:selective_participation_effect} visualizes this comparison, with the same ten-percentage-point vertical span in all three panels.
The lines connect the two aggregate means, while the error bars indicate sample standard deviations across training runs.
The thermal and event improvements are observed in all three matched repetitions, whereas the depth improvement is positive in the mean but not in every repetition.
The strongest evidence for the selective rule therefore comes from the thermal and event comparisons, with a smaller and more variable depth effect.

\begin{figure*}[!t]
  \centering
  \includegraphics[width=.98\textwidth]{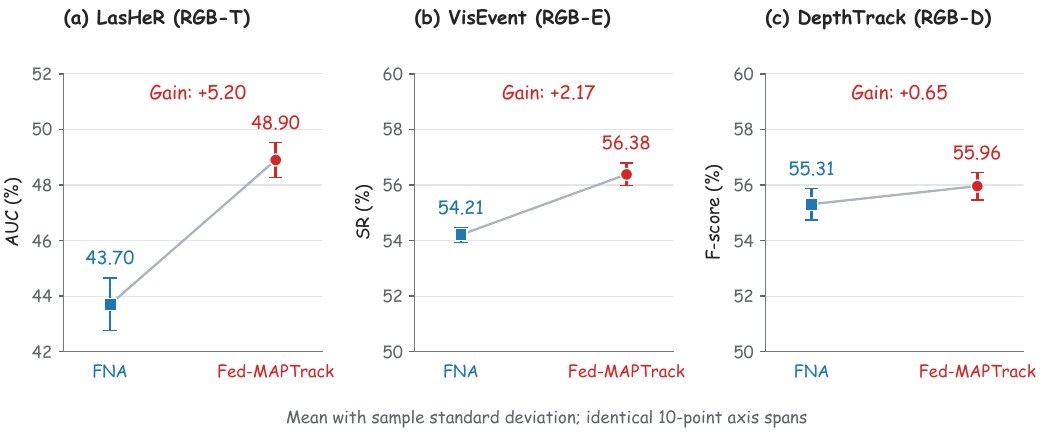}
  \caption{Effect of selective participation with the parameter-group bucketing scope held fixed. Blue squares and red circles show the mean scores of FNA and \methodname, respectively, while error bars indicate sample standard deviations across three training repetitions. The mean gains are 5.20 AUC points on LasHeR, 2.17 SR points on VisEvent, and 0.65 F-score points on DepthTrack. Each vertical axis spans ten percentage points. Connecting lines indicate changes between aggregate means, rather than individual-run trajectories.}
  \label{fig:selective_participation_effect}
\end{figure*}

\textbf{Interpreting the complete design.}
Compared with auxiliary-only bucketing, the complete method increases mean thermal AUC by 1.43 percentage points, while its event and depth means are slightly lower by 0.08 and 0.06 percentage points.
These differences do not support uniform improvement over auxiliary-only bucketing across all modalities.
The ablation instead shows that the extended sharing structure requires a compatible participation rule: S-A recovers the performance lost under FNA and provides the clearest additional benefit in thermal tracking.
Bucketing defines where modality-related information is shared, while S-A determines which clients may contribute to each sharing domain.
The two mechanisms form a combined aggregation strategy rather than two competing final methods.

\subsection{Selective Participation Analysis}
\label{subsec:rgb_only_analysis}

To address RQ3, we examine the data available to RGB-only clients and then analyze the final parameters of FNA and \methodname.
Client composition provides the context for sample-weighted aggregation, while the parameter analyses describe which groups differ after training.
The controlled ablation establishes the performance differences. The following analyses examine data allocation and parameter patterns that may help explain them.

\textbf{Heterogeneous data allocation.}
Figure~\ref{fig:rgb_only_composition} shows the training-sample composition of the evaluated client partition.
LasHeR contains 238,827 training samples assigned to sensor-equipped clients and 220,295 assigned to RGB-only clients. The corresponding counts are 32,072 and 179,990 for VisEvent and 203,709 and 6,204 for DepthTrack.
The corresponding RGB-only proportions are 48.0\%, 84.9\%, and 3.0\%.
Although each benchmark supplies exactly one RGB-only client among four clients, the fraction of data assigned to that client differs substantially.
Thus, an identical client-count ratio does not imply an identical sample-weighted contribution to aggregation.
These counts describe the fixed training partition, rather than the number of evaluated frames or newly acquired RGB-only sequences.

\begin{figure}[pos=t]
  \centering
  \includegraphics[width=\columnwidth]{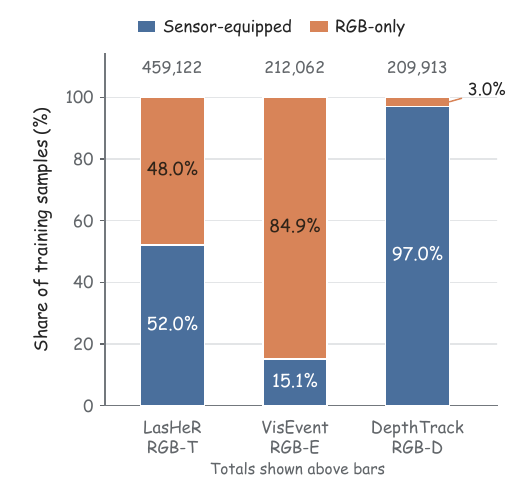}
  \caption{Training-sample composition of sensor-equipped and RGB-only clients under the evaluated partition. Totals are shown above the bars. RGB-only samples account for 48.0\%, 84.9\%, and 3.0\% of LasHeR, VisEvent, and DepthTrack, respectively. Each benchmark has three sensor-equipped clients and one RGB-only client, while the bars show the proportions of training samples assigned to these two client types.}
  \label{fig:rgb_only_composition}
\end{figure}

\textbf{Participation at the parameter-group level.}
The relevant question is which parameter groups an RGB-only client should help aggregate, rather than whether that client should be removed from federated training.
FNA and \methodname\ retain the same clients and assigned RGB data.
In both configurations, RGB-only clients perform local training, receive the fusion and normalization banks associated with their originating dataset, and contribute their RGB-shared updates to the global bank.
S-A changes only the inclusion of their fusion and normalization updates in the server's modality-specific averages.
The resulting FNA-to-\methodname\ comparison therefore isolates aggregation eligibility without deleting clients, reducing the local RGB training budget, or replacing their assigned data.

The thermal and event gains support choosing contributors according to the role of the parameter group rather than treating RGB-only clients as universally unsuitable collaborators.
Their RGB-shared contributions remain present in both configurations.
The present ablation isolates eligibility for fusion and normalization aggregation. Quantifying the benefit of RGB-only clients' contributions to the shared RGB bank would require a separate participation ablation.

\textbf{Interpreting the composition and performance together.}
The composition figure provides context for the participation experiment but is not a controlled sweep of the RGB-only ratio.
VisEvent has the largest RGB-only data proportion, whereas the largest FNA-to-\methodname\ mean gain occurs on LasHeR.
The ratio alone therefore does not explain the magnitude of the performance change.
The benchmarks also differ in auxiliary modality, sequence distribution, and within-benchmark data imbalance.
We interpret the combined evidence as support for parameter-group eligibility under the current partition, without inferring a universal exclusion threshold or a monotonic relation between RGB-only proportion and tracking improvement.

\textbf{Final weight distributions across parameter groups.}
Figure~\ref{fig:prompt_weight_distributions} compares the empirical final weight distributions of FNA and \methodname\ at round 120.
RGB-shared weights come from the global bank, while auxiliary, fusion, and normalization weights are pooled across the active thermal, event, and depth banks.
The analysis includes only weight tensors, excluding biases and scalar fovea parameters. For normalization, these weights are the scale parameters.
Within each group, both methods use the same 100 histogram bins covering the complete observed range.
Per-run densities are normalized and then averaged across the three repetitions.

\begin{figure}[pos=t]
  \centering
  \includegraphics[width=\columnwidth]{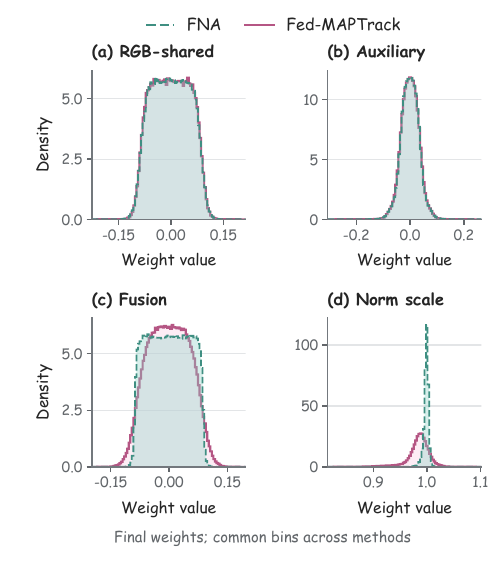}
  \caption{Empirical final weight distributions of FNA and \methodname\ at communication round 120. RGB-shared weights come from the global bank, while the other groups pool weights from the three active modality banks. Curves average three per-run densities using common histogram bins covering the full observed range. Biases, scalar fovea parameters, stored initial banks, and inactive global modality banks are excluded. Panel (d) shows normalization scale weights. These are final-state distributions, not changes from initialization.}
  \label{fig:prompt_weight_distributions}
\end{figure}

The RGB-shared and auxiliary weight distributions largely overlap, whereas differences are more visible in fusion and normalization.
In particular, FNA normalization scales are concentrated close to one, while the complete method has a broader distribution centered below one.
The final distributional differences are therefore parameter-group dependent and most apparent in groups whose RGB-only participation rule changes.
Marginal histograms do not preserve parameter identity, so overlapping distributions do not imply identical tensors and broader distributions do not establish better adaptation.
The next analysis retains the correspondence between the parameters of different modality banks.

\textbf{Layer-wise separation of modality banks.}
To preserve the identity of corresponding parameters across modalities, Figure~\ref{fig:prompt_bank_dispersion} examines the thermal, event, and depth banks within each of the twelve Transformer blocks.
For parameter group $g$ and block $\ell$, let $\boldsymbol{\theta}^{m}_{g,\ell}$ concatenate the corresponding parameters, including biases, in modality bank $m$.
Their element-wise bank mean is $\bar{\boldsymbol{\theta}}_{g,\ell}$, with $\mathcal{M}=\{T,E,D\}$.
We compute relative bank dispersion as
\begin{equation}
  D_{g,\ell}=100\sqrt{
    \frac{\sum_{m\in\mathcal{M}}
      \|\boldsymbol{\theta}^{m}_{g,\ell}-\bar{\boldsymbol{\theta}}_{g,\ell}\|_2^2}
    {\sum_{m\in\mathcal{M}}
      \|\boldsymbol{\theta}^{m}_{g,\ell}\|_2^2}
  }.
  \label{eq:bank_dispersion}
\end{equation}
The mean is taken element-wise over the three modality banks.
The statistic is their RMS deviation from this mean divided by the pooled RMS parameter magnitude, expressed as a percentage.
Corresponding tensor names and shapes are aligned before concatenation.
Each repetition is analyzed separately, followed by the mean and sample standard deviation across repetitions.
The analysis includes only active final modality banks, excluding stored initial states and inactive global modality banks.
The RGB-shared group is omitted because it has only one global bank.

\begin{figure*}[!t]
  \centering
  \includegraphics[width=\textwidth]{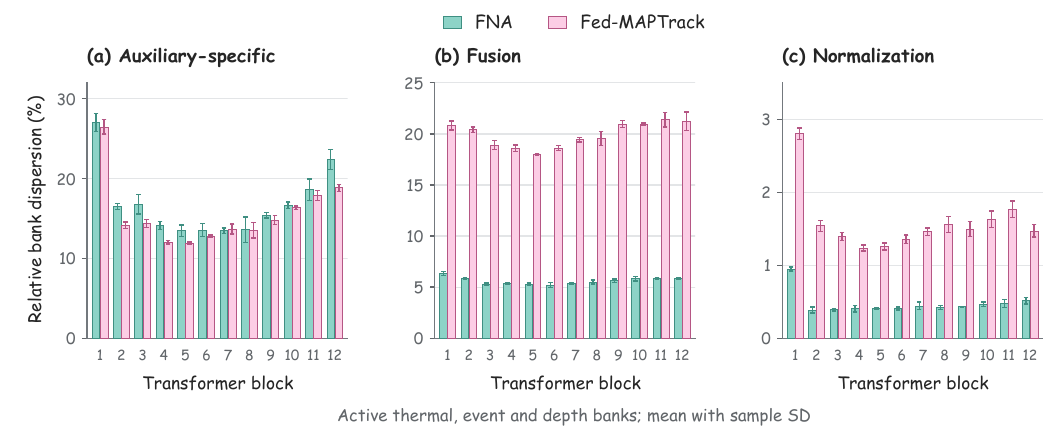}
  \caption{Layer-wise separation of final thermal, event, and depth parameter banks, measured by the relative dispersion in Eq.~\ref{eq:bank_dispersion}. All parameters, including biases, are included within each group and Transformer block. Bars and error bars show the mean and sample standard deviation across three training repetitions. The RGB-shared group has only one bank and is omitted. Each panel starts at zero but uses its own labeled range. The statistic describes final bank separation, not tracking accuracy or changes from initialization.}
  \label{fig:prompt_bank_dispersion}
\end{figure*}

The complete method has higher fusion and normalization bank dispersion than FNA in every displayed block.
For fusion, mean dispersion ranges from approximately 18.0\% to 21.4\%, compared with 5.2\% to 6.3\% for FNA.
For normalization, the corresponding ranges are 1.24\%--2.80\% and 0.39\%--0.95\%.
Auxiliary-specific dispersion does not increase uniformly and is lower for the complete method in most blocks.
Thus, the observed differences are not a uniform increase in separation across every parameter group.
They are concentrated in fusion and normalization, consistent with the design goal of preserving modality-dependent adaptation in groups that interact with auxiliary observations.

Taken together, the ablation and parameter analyses support a coherent interpretation: sharing domains alone do not determine an effective aggregation rule, and contributor eligibility matters particularly for fusion and normalization.
The complete method improves performance over FNA while exhibiting more distinct final modality banks in these groups.
This correspondence supports the interpretation of group-specific adaptation, but separation itself is not a quality metric or proof that it causes the tracking gain.
Both figures describe final parameter states and do not measure changes from initialization or gradient conflict between clients.

\subsection{Optimization and Tracking Analysis}
\label{subsec:qualitative_efficiency}

\textbf{Recorded local optimization.}
To address the optimization part of RQ4, Figure~\ref{fig:federated_training_loss} uses the existing per-client loss logs over all 120 communication rounds.
Within each repetition and round, we average the twelve clients' mean local training losses with equal client weights.
We then calculate the mean and sample standard deviation across the three repetitions.
All recorded rounds are displayed without smoothing or interpolation.
This diagnostic uses equal client weights and is therefore distinct from the sample-weighted server aggregation.

\begin{figure}[pos=t]
  \centering
  \includegraphics[width=\columnwidth]{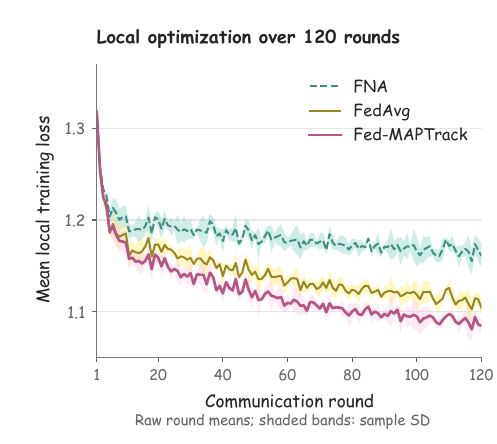}
  \caption{Recorded local training loss over 120 communication rounds. Within each repetition, the twelve clients' mean local losses are averaged equally in each round. Curves and shaded bands then show the mean and sample standard deviation across three repetitions. All recorded rounds are plotted without smoothing or interpolation. The figure describes local optimization, not intermediate-round benchmark performance.}
  \label{fig:federated_training_loss}
\end{figure}

The complete method attains lower recorded mean local losses than FedAvg and FNA over much of training, while all three show an overall declining trajectory.
This is consistent with the aggregation design supporting local optimization under the current client objectives.
Because the values are collected during local training, they are not losses of one server model evaluated on a common held-out set.
Moreover, the available benchmark evaluations concern round 120 only.
We therefore do not interpret the curves as test-performance convergence, time to a target accuracy, or evidence of improved generalization by themselves.

\textbf{Qualitative tracking comparison.}
To address the localization part of RQ4, Figure~\ref{fig:qualitative_results} compares saved predictions of FedAvg, Independent FL, and \methodname\ on sequences from the three benchmarks.
The displayed sequences are basketboy from LasHeR, personFootball from VisEvent, and backpack\_indoor from DepthTrack.
Each row presents temporally ordered frames from one sequence, and all methods use the same reference image and crop within each comparison.
The enlarged insets contain the ground truth and all displayed predictions, allowing differences in location and target extent to be inspected under a common spatial scale.
The examples come from one fixed training repetition and supplement, rather than replace, the repeated benchmark evaluations.

\begin{figure*}[!t]
  \centering
  \includegraphics[width=\textwidth]{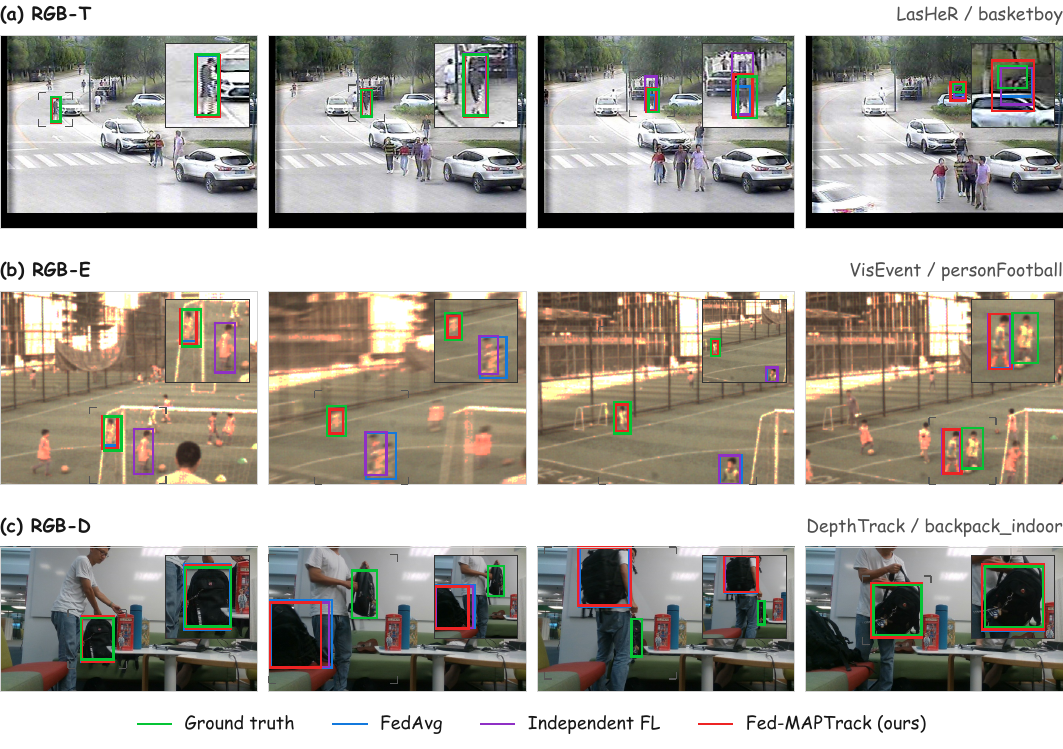}
  \caption{Qualitative RGB-X tracking comparisons on (a) LasHeR, (b) VisEvent, and (c) DepthTrack. Green, blue, purple, and red solid boxes indicate ground truth, FedAvg, Independent FL, and \methodname, respectively. Each row shows temporally ordered frames from one sequence. Insets magnify a common region containing all displayed predictions and the ground truth, with gray corner markers identifying the corresponding regions in the full images. The examples include both accurate localization and tracking failures.}
  \label{fig:qualitative_results}
\end{figure*}

In the event sequence, intermediate examples show \methodname\ remaining near the annotated target while the compared methods localize another person.
The thermal sequence illustrates differences in target location and predicted extent, together with a later example in which the complete method also drifts.
The depth sequence includes difficult frames in which the displayed trackers deviate from the annotated target.
These examples show both the potential benefit and the remaining failure modes of the aggregation strategy.
They do not establish a benchmark-wide recovery rate or a guarantee of robustness to a particular tracking attribute.
The quantitative comparisons remain those reported in Tables~\ref{tab:main_results} and~\ref{tab:ablation}.

\textbf{Box-derived localization comparison.}
Figure~\ref{fig:box_localization} provides a closer comparison between FNA and \methodname\ on basketboy.
For each method, the displayed localization map is constructed from its saved predicted bounding box using the same elliptical Gaussian encoding.
The Gaussian is centered at the predicted box center, and its horizontal and vertical standard deviations are one quarter of the predicted box width and height.
Both methods use the same crop coordinates and color scale, with ground-truth boxes overlaid for spatial reference.
This makes differences in predicted location and extent visible without altering the saved box predictions.

\begin{figure}[pos=t]
  \centering
  \includegraphics[width=\columnwidth]{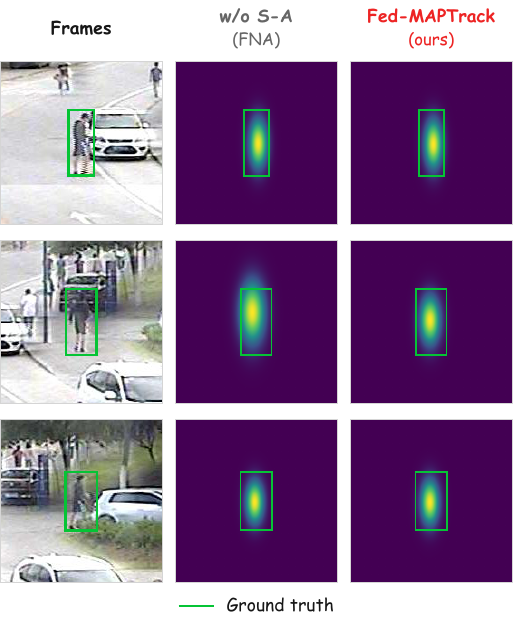}
  \caption{Box-derived localization comparison between FNA and \methodname\ on basketboy. The first column shows reference RGB crops, while the remaining columns visualize saved box predictions using the same elliptical Gaussian encoding. Each Gaussian is centered on the predicted box, with horizontal and vertical standard deviations equal to one quarter of its width and height, respectively. Green boxes indicate ground truth in the common crop coordinates. }
  \label{fig:box_localization}
\end{figure}

The middle displayed example illustrates a difference in estimated target extent between FNA and the complete method, whereas other examples show much closer predictions.
Because the maps are derived analytically from predicted boxes, their spatial concentration follows the box geometry.
They are not measured attention, activation, or confidence maps and are not used as evidence for improved internal attention concentration.
Their role is to complement the tracking-box comparison with an explicit visualization of localization outputs.

\subsection{Communication Analysis}
\label{subsec:communication}

During federated optimization, clients exchange only prompt-related parameters, while the frozen backbone and prediction head remain local.
The experiments use a local federated simulation rather than a measured network deployment, so we account for FP32 tensor payloads explicitly.
A full local prompt view contains 839,628 parameters, corresponding to 3,358,512 bytes.
The client implementation constructs prompt-state dictionaries for all four groups and uses a separate validity mask to identify eligible updates.
An RGB-only client's valid RGB-shared, fusion, and normalization groups contain 175,212 parameters, or 700,848 bytes, but marking its auxiliary tensors invalid does not itself remove them from that dictionary.

If full prompt-state dictionaries are serialized in both directions for all twelve clients, the tensor-only payload is 80.60 MB per round and 9.673 GB over 120 rounds.
If a transport implementation additionally removes invalid RGB-only auxiliary tensors before upload while retaining full prompt downloads, the corresponding estimate becomes 72.63 MB per round and 8.716 GB over 120 rounds.
The latter is a mask-aware transport estimate, not measured traffic or an automatic consequence of the aggregation mask.
Both calculations use decimal units and exclude serialization metadata and network protocol overhead.

The core aggregation variants share the same local prompt-state dimensions.
S-A filters RGB-only fusion and normalization updates at the server rather than removing those tensors from the client state.
Bucketing defines sharing domains, while S-A determines eligible contributors. Neither mechanism provides additional communication compression.
The smaller parameter payload relative to full-model exchange comes from prompt tuning, whereas the controlled comparisons evaluate the aggregation design.
This accounting does not establish latency, energy, wall-clock training, or deployment improvements.

\par
\endgroup

\section{Discussion and Limitations}
\label{sec:discussion}

\textbf{Joint design of bucketing and selective participation.}
The central finding is that the sharing domain and contributor eligibility must be considered together.
Extending bucketing to fusion and normalization without S-A reduces performance, whereas adding S-A improves FNA and yields higher means than FedAvg on all three evaluated benchmarks.
The complete method also achieves higher mean scores than Local, FedProx, FedPer, Ditto, FedMSplit-P, and FED-PRIME on all three benchmarks, while Independent FL has a slightly higher event mean.
Compared with auxiliary-only bucketing, the complete method provides the clearest additional benefit on LasHeR, while its event and depth means are slightly lower by less than 0.1 points.
Final-parameter analyses reveal corresponding group-dependent differences, but do not establish a causal relation between bank separation and accuracy.

\textbf{Evaluation scope.}
The experiments use twelve simulated clients and three training repetitions under one fixed partition.
Modality, dataset, and RGB-only data proportion vary together, so the present results do not isolate a universal effect of the RGB-only ratio or establish generality across partitions.
Communication figures are tensor-payload calculations rather than measured edge-device latency or energy savings.
Keeping raw videos local does not by itself prevent information leakage through exchanged parameters. Formal privacy protection is beyond the scope of this evaluation.

\textbf{Dynamic sensor conditions.}
S-A uses client-level sensor availability rather than a learned estimate of auxiliary observation quality.
Sample-level masking is included in training, but noisy, delayed, or persistently failing sensors may require finer eligibility decisions.
Observation-quality estimation and dynamic sensor availability are therefore natural directions for extending the participation rule.

\section{Conclusion and Future Work}
\label{sec:conclusion}

We presented \methodname, a federated RGB-X tracking framework for clients with heterogeneous auxiliary sensors and RGB-only observations.
The framework adapts prompt-related parameters on a frozen backbone and prediction head, keeping raw training videos local.
Its unified aggregation strategy shares RGB parameters globally and combines modality-specific auxiliary, fusion, and normalization banks with group-specific contributor eligibility.
RGB-only clients retain their shared RGB contributions, while their fusion and normalization updates are excluded from modality-specific server averages.
Experiments on LasHeR, VisEvent, and DepthTrack show that \methodname\ achieves higher mean scores than FedAvg under the evaluated partition.
The complete method also achieves higher mean scores than Local, FedProx, FedPer, Ditto, FedMSplit-P, and FED-PRIME on each benchmark.
It attains the highest thermal AUC and depth F-score among the evaluated methods, while its event SR remains 0.07 points below Independent FL.
Controlled ablations show that extending the bucketing scope alone is insufficient and that a compatible participation rule is essential to the complete design.

Future work will examine observation-quality-aware participation, dynamic sensor availability, and finer parameter-sharing rules.
Evaluation across additional partitions and data proportions, together with measured edge-device communication and runtime, will further assess generality and deployment suitability.

\printcredits

\bibliographystyle{cas-model2-names}
\bibliography{cas-refs}

\end{document}